\UseRawInputEncoding
\documentclass[letterpaper, 10 pt, conference]{ieeetran}
\IEEEoverridecommandlockouts
\usepackage{graphicx} 
\usepackage{float} 
\usepackage{subfigure} 
\usepackage{cite}
\usepackage{amssymb}
\usepackage{amsmath}
\usepackage{booktabs}
\usepackage{threeparttable}
\usepackage{makecell}
\usepackage{balance}
\usepackage[colorlinks,linkcolor=black]{hyperref}
\IEEEoverridecommandlockouts 
\title{\LARGE \bf
Reinforcement Learning Based Pushing and Grasping 

Objects from Ungraspable Poses
}
\author{Hao Zhang$^{1,2,3}$, Hongzhuo Liang$^{3}$, Lin Cong$^{3}$, Jianzhi Lyu$^{3}$, Long Zeng$^{1*}$, Pingfa Feng$^{1}$, and Jianwei Zhang$^{3}$ 
\thanks{This work was done when Hao Zhang was visiting Universit\"{a}t Hamburg.}
\thanks{This research was funded by the German Research Foundation (DFG)
	and the National Science Foundation of China (NSFC)
	in project Crossmodal Learning DFG TRR-169/NSFC 62061136001, the National Natural Science Foundation of China (Grant No. 61972220), Guangdong Natural Science Fund-General Programme Grand No. 2022A1515011234
	and partially supported by European project H2020 Ultracept under Grant 778602.}
\thanks{$^{*}$ Corresponding author, email: zenglong@sz.tsinghua.edu.cn}%
\thanks{$^{1}$ Division of Advanced Manufacturing, Shenzhen International Graduate School, Tsinghua University.}%
\thanks{$^{2}$ Production Systems Engineering, RWTH Aachen University.}%
\thanks{$^{3}$ Group TAMS, Dept. of Informatics, Universit\"{a}t Hamburg.}%
}

\begin{document}
\maketitle
\thispagestyle{empty}
\pagestyle{empty}

\begin{abstract}
Grasping an object when it is in an ungraspable pose is a challenging task, such as books or other large flat objects placed horizontally on a table. Inspired by human manipulation, we address this problem by pushing the object to the edge of the table and then grasping it from the hanging part. In this paper, we develop a model-free Deep Reinforcement Learning framework to synergize pushing and grasping actions. We first pre-train a Variational Autoencoder to extract high-dimensional features of input scenario images. One Proximal Policy Optimization algorithm with the common reward and sharing layers of Actor-Critic is employed to learn both pushing and grasping actions with high data efficiency. Experiments show that our one network policy can converge 2.5 times faster than the policy using two parallel networks. Moreover, the experiments on unseen objects show that our policy can generalize to the challenging case of objects with curved surfaces and off-center irregularly shaped objects. Lastly, our policy can be transferred to a real robot without fine-tuning by using CycleGAN for domain adaption and outperforms the push-to-wall baseline.
\end{abstract}

\section{INTRODUCTION}
\begin{figure}[t] 
	\centering 
	\includegraphics[width=7.5cm]{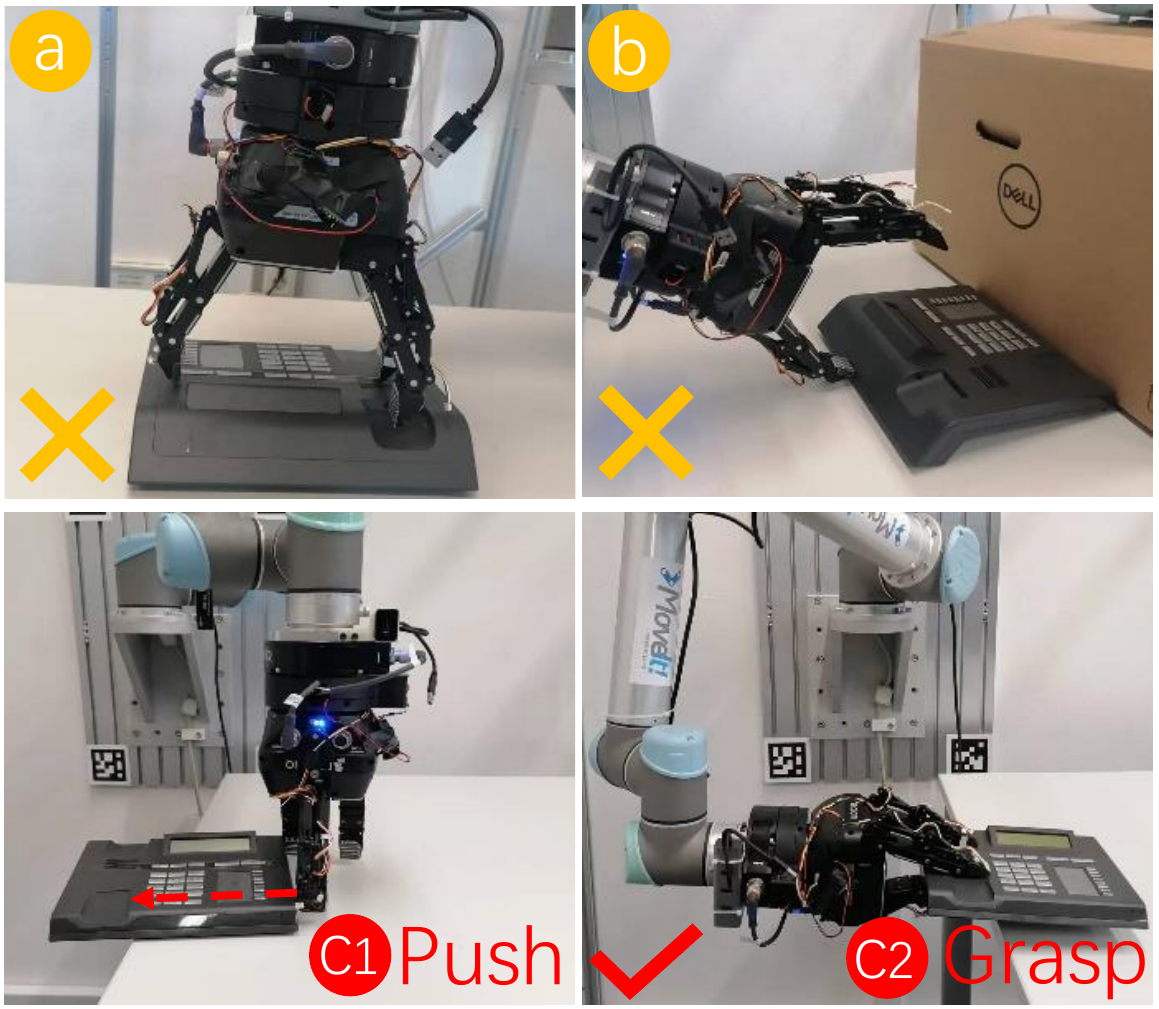} 
	\caption{Demonstrates previous failure cases and our push-and-grasp (PaG) method for ungraspable posed objects. Our PaG method consists of two parts: (1) pushing the object to the edge of the table; (2) grasping the hanging part of the object on the edge of the table.} 
	\label{Fig. 1} 
\end{figure}

Imagine a landline phone or other large flat objects placed on a table with only its height less than the max stroke of the gripper. There is no graspable position existing in this configuration, as shown in Fig.~\ref{Fig. 1}(a). For the human grasping strategy, we intend to first make it in a semi-suspend position and then grasp it from the side. If there is a wall near the object, we can push it against the wall to facilitate grasping~\cite{liang2021shovel_and_grasp, sun2020push_to_wall}. However, this method restricts a wall in the environment and requires that the object has no bevels on the side being pushed. Otherwise, it will fail, as shown in Fig.~\ref{Fig. 1}(b). Instead of pushing it to a wall, we can also push it to the edge of the table and then grasp it from the hanging part, as shown in Fig.~\ref{Fig. 1}(c1-c2). The push-to-edge approach is insensitive to the surrounding environment as well as the shapes and materials of objects.

Recently, visual pushing and grasping~\cite{zeng2018vpg} and push-to-wall~\cite{liang2021shovel_and_grasp, sun2020push_to_wall} have been proposed to use push as a pre-grasp manipulation to create graspable poses. Visual pushing and grasping refers to the use of pushing to separate objects in a stacked scenario, while push-to-wall focuses on using push to make the large object to the wall in a semi-hanging state. Although they synergize pushing and grasping manipulation to help accomplish their tasks, there are still some limitations. Visual pushing and grasping using two parallel Deep Q-Networks (DQN)~\cite{mnih2013dqn} predicts actions leading to data inefficiency and coordination problems of the two actions. Push-to-wall does not use an end-to-end approach, it employs a Deep Reinforcement Learning (DRL) policy to learn pushing actions and a hard-coded method to execute grasping actions, which weakens the synergy between pushing and grasping. At the same time, they are also restricted by the shapes and materials of the objects which can not handle the beveled objects or smooth objects.

It is a challenging task that efficiently learns an end-to-end policy of pushing the large flat objects to the edge of the table and then grasping from the hanging part. The policy should learn the trade-off between pushing the object to an easy-to-grasp position and avoiding falling to the ground. In addition, the policy should select appropriate grasping positions according to the object pose to achieve successful grasping. The traditional motion planning method is difficult to handle the random position of the objects as well as the sliding and dynamic situations of objects during the pushing and grasping process. We use DRL in our method to achieve better control of the actions, and we only employ one policy network to reuse the observations and reduce parameters for higher data efficiency.

This paper proposes a model-free DRL framework called PaG for visual-based pushing and grasping manipulations. We pre-train a Variational Autoencoder (VAE)~\cite{kingma2013vae} to extract high-dimensional features of the RGB images of the scene. Then, we concatenate the features extracted from images with gripper pose as the state of a Markov Decision Process (MDP). We set the distance between the gripper and the object along the x-direction as a common reward for both the pushing and grasping stages (see Fig.~\ref{Fig. 2}) because we want the robot to push stably along the center of the object and grasp also close to the center of the object. Through common reward and sharing layers of Actor-Critic, we can guide the policy to learn pushing and grasping as a whole process with high data efficiency by using one Proximal Policy Optimization (PPO)~\cite{schulman2017ppo} model. We adopt CycleGAN to achieve transfer to a real robot without fine-tuning.

The main contributions of this paper are as follows: 
\begin{itemize}
\item We introduce an end-to-end reinforcement learning framework to tackle the challenging robotic manipulation task of grasping large flat objects from ungraspable poses based on vision.
\item We achieve high data efficiency by employing one PPO network to train pushing and grasping actions through a common reward and sharing layers of Actor-Critic.
\item We demonstrate the generalization capability of our policy using beveled objects and irregular objects with non-overlapping centers of mass and shape. Moreover, the transfer from simulation to a real robot can be achieved without fine-tuning.
\end{itemize}

\section{RELATED WORKS}

\subsection{Synergy of Pushing and Grasping}
The synergy of pushing and grasping actions is introduced to grasp objects under restricted scenarios and to improve the quality of grasping by applying pushing actions. Zeng \textit{et al.}~\cite{zeng2018vpg} proposed using the action of pushing to rearrange objects in a clustered scene to make space for grasping. Xu \textit{et al.}~\cite{xu2021grasptarget} introduced a goal-oriented grasping framework to achieve stable grasping of the target object in a cluttered environment by pushing away interfering objects. 

The above methods can handle some challenging stacked scenarios but only focus on small blocks. Liang \textit{et al.}~\cite{liang2021shovel_and_grasp} and Sun \textit{et al.}~\cite{sun2020push_to_wall} both expanded methods for grasping the large flat object by first pushing it against a wall to make it in a semi-suspended state and then grasping from the side. The former only used DRL to predict the pushing action and hard-code the grasping action after each pushing step, while the latter used DRL to push the object against the wall and then grasp it from the side using another robot through motion planning. In contrast, our PaG method employs DRL synergy in both pushing and grasping actions to achieve better end-to-end pushing and grasping of large flat objects and does not require a wall in the environment.

\subsection{Reinforcement Learning for Robotic Manipulation}
Reinforcement learning is a general framework for learning skills from scratch through interaction with the environment, guided by maximizing the reward feedback from the environment. Cong \textit{et al.}~\cite{cong2022reinforcement} used a Soft Actor-Critic~\cite{haarnoja2018sac} network to push object to target position. Breyer \textit{et al.}~\cite{breyer2019closeloopgrasp} devised a framework depending on Trust Region Policy Optimization~\cite{schulman2015trpo} to pick objects. Zeng \textit{et al.}~\cite{zeng2018vpg} and Yang \textit{et al.}~\cite{yang2021pushandgrasp} composed two parallel Deep Q-Networks~\cite{mnih2013dqn} to push and grasp objects in stacked scenarios. Unlike previous work, which always required two parallel networks for both pushing and grasping actions, our method uses only one network to train the pushing and grasping actions with higher data efficiency.

\begin{figure*}[t]
	\centering
	\includegraphics[width=17cm]{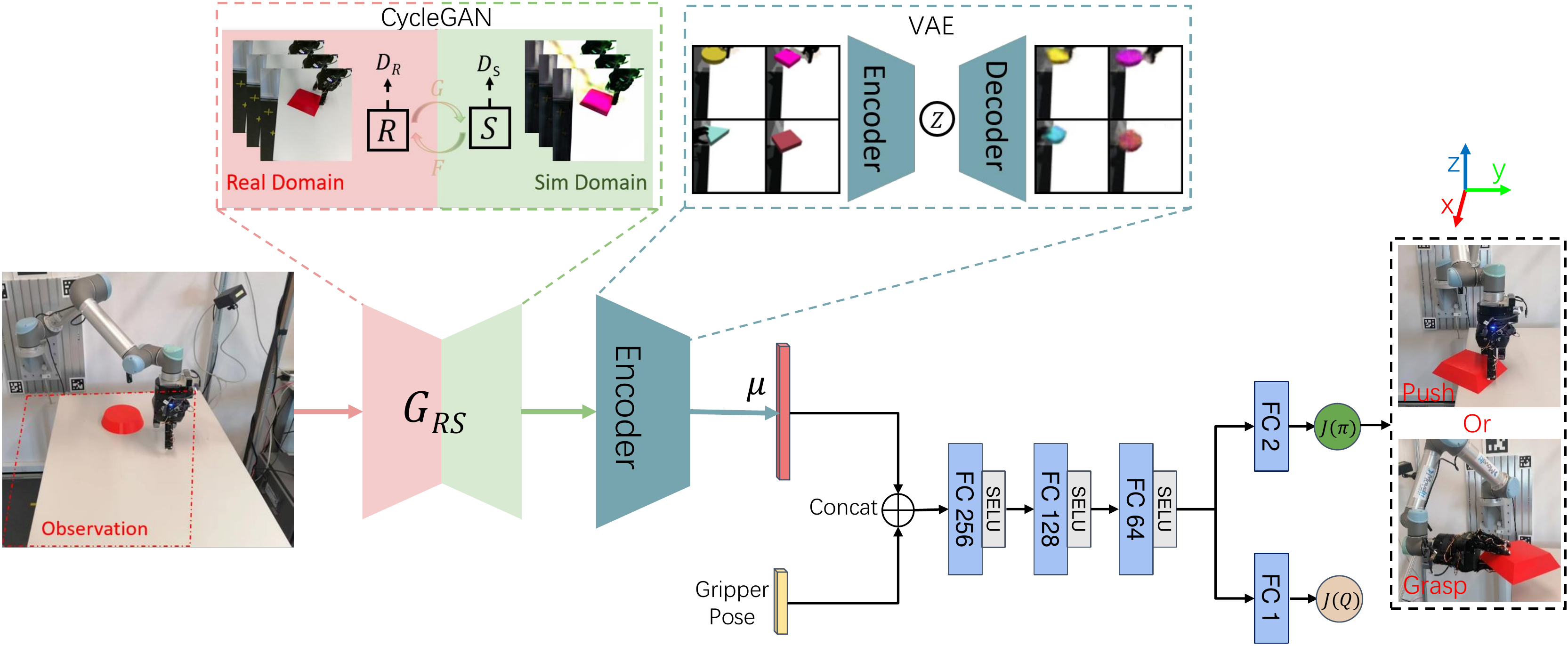}
	\caption{Overview of our PaG method. We pre-train a VAE for feature extraction and a CycleGAN for domain adaption. We train our policy in Isaac Gym simulator, and the pre-trained encoder obtains a latent vector concatenated with the gripper pose as a state embedded into the policy network (shared layers of Actor-Critic) to determine the following action (144 steps for push and 84 steps for grasp). For the real robot experiment, the pre-trained CycleGAN transfers the images from the real domain to the simulation domain, effectively tricking the policy to believe that it is still in the simulator.} 
	\label{Fig. 2}
\end{figure*}

\subsection{Sim2Real}
Thanks to the development of physical simulators such as PyBullet~\cite{coumans2016pybullet}, MuJoCo~\cite{todorov2012mujoco}, and Isaac Gym~\cite{makoviychuk2021isaacgym}, we can use reinforcement learning methods to make the robot learn faster and safer in the simulator and then transfer it to a real robot. However, deploying the policy trained in simulation on a real robot requires bridging the gap between the two domains, especially when using RGB images as observations. One approach is to use the domain randomization method to train a more robust policy that randomly changes color, texture, lighting, and camera position~\cite{tobin2017sim2real, james2017sim2real}. Another approach employs system identification to find the correct simulation parameters making the simulation match the real systems as close as possible~\cite{tan2018sim2real}. In our work, we use CycleGAN~\cite{zhu2017cyclegan} to transfer realistic images to a simulated style when applying the trained model to a real robot.

\section{METHOD}
In this section, we will present our visual-based framework for pushing and grasping manipulations. This includes visual representation using VAE, reinforcement learning policy employing PPO, and sim2real transfer applying CycleGAN.

\subsection{Visual Representation}
\label{Visual Representation}
We get the task-relevant compact representations from the input images by using VAE~\cite{kingma2013vae}, a probabilistic generative model consisting of an encoder that converts the input image to a prior distribution $q_{\theta}(z|x)$ and a decoder that converts the latent vector $z$ into a distribution $p_{\omega}(x|z)$, as depicted in Fig.~\ref{Fig. 2}. The input image is a $64 \times 64 \times 3$ RGB image from the simulator scenario. The encoder consists of five $3 \times 3$ convolutional layers to obtain a 128-dimensional latent vector. The first five layers of the decoder mirror the structure of the encoder, and there is an additional $3 \times 3$ convolutional layer with stride 1 to avoid checkerboard artifacts. The model is trained by minimizing the reconstruction loss between input and output images and forcing the latent representation to a prior distribution. Therefore, the loss function can be described as follows:
\begin{equation}
L(\theta,\omega) = -\mathbb{E}_{q_{\theta}(z|x)}[{\rm log}p_\omega(x|z)]+\mathbb{KL}(q_{\theta}(z|x)||p(z))
\end{equation}

The VAE part in Fig.~\ref{Fig. 2} shows the examples of input and output images trained on a dataset of 100000 images from the simulator, using 165 epochs of the Adam optimizer~\cite{kingma2014adam}, with a learning rate of 0.005.

\subsection{Reinforcement Learning}
A large flat object on the table requires a series of pushing actions to move to the edge of the table and then grasp from the hanging part. This sequential decision-making problem can be formulated as a long-horizon MDP.

\subsubsection{\itshape State}
The state of the DRL agent consists of two parts, one is a latent vector representing the corresponding position information between the gripper, object, and table, and the other is the proprioceptive state of the robot. For the first part, we map the simulator image $I_s$ to a 128-dimensional latent vector $z$ by employing the pre-trained encoder in Section \ref{Visual Representation}. The latent vector $z$ samples form Gaussian distribution $q_{\theta}(z|I_s) =  \mathcal{N}(\mu_{\theta}(I_s), \sigma_{\theta}^2(I_s))$, we take the mean of the encoder $\mu_{\theta}(I_s)$ as the state representation. For the second part, we adopt the 7-dimensional vector gripper pose $G_p$ to characterize the robot state. Therefore, the state can be formed as:
\begin{equation}
S = [\mu_{\theta}(I_s), G_p]
\end{equation}

\subsubsection{\itshape Action}
\label{Action}
The agent $\pi(s_t)$ outputs incremental actions:
\begin{equation}
A = [\Delta p_x, \Delta p_y, \Delta g_x]
\end{equation}
where the actions bound are from -0.01$m$ to 0.01$m$. We divide the whole process into four stages by time steps. The first 144 steps push the large flat object to the edge of the table by using incremental actions $\Delta p_x, \Delta p_y$. The next 48 steps use motion planning to move the robot from the end of the pushing pose to the predefined start grasping pose. The next 84 steps find a proper grasping position by using incremental actions $\Delta g_x$. In the final 44 steps, close the gripper and lift the object. For the first (pushing) and third (grasping) stages, we want the gripper and the object to be in stable contact, which means we want the gripper and the object to have the same absolute coordinates in the x-direction for both stages, as shown in Fig.~\ref{Fig. 2}. Based on the same purpose of both stages, we equal $\Delta p_x$ and $\Delta g_x$ to simplify the action space.

\subsubsection{\itshape Reward}
\label{Reward}
We adopt pushing, grasping, and common rewards to guide policy learning the long horizon pushing and grasping actions. $R_p = 1$ for pushing to an appropriate position that is convenient for grasping and avoiding falling (the absolute coordinates of the object in y-direction $o_y$, $0.85<o_y<0.9$). $R_p = -1$ for pushing to a dangerous position where the object is about to fall ($o_y<0.84$). $R_p = -1$ for object falls down (the absolute coordinates of the object in z-direction $o_z$, $o_z<0.7$). $R_g = 5$ for successful grasping ($o_z>0.83$). We adopt a common reward $R_c$ for stable contact between the object and the gripper during the pushing and grasping stages (see Sec. \ref{Action}), which is proportional to the distance difference between the object and the gripper in the x-direction ($R_c = -2.5\mid o_x-G_{p_x}\mid$). Overall, the reward can be formed as:
\begin{equation}
	R = R_p+R_g+R_c
\end{equation}

\subsubsection{\itshape Policy optimization}
\label{Policy optimization}
We employ one PPO policy to learn pushing and grasping actions with high data efficiency by using a common reward (see Sec. \ref{Reward}) and sharing layers of Actor-Critic (see Fig.~\ref{Fig. 2}). The common reward is inspired by multi-task reinforcement learning~\cite{yu2020meta}. In our scenario, pushing and grasping can be regarded as two related tasks since both successful pushing and grasping require the gripper to be in stable contact with the object, which means they share some common structures that can be represented by a common reward. The common reward can guide the agent to learn pushing and grasping actions as a whole process that we can learn with one PPO model. We can share non-output layers of Actor-Critic for higher data efficiency as some related works~\cite{mnih2016a3c, shukla2021shareac}. The relevant experiments can be found in Section \ref{Policy Training}.

\begin{figure}[t] 
	\centering 
	\includegraphics[width=7.5cm]{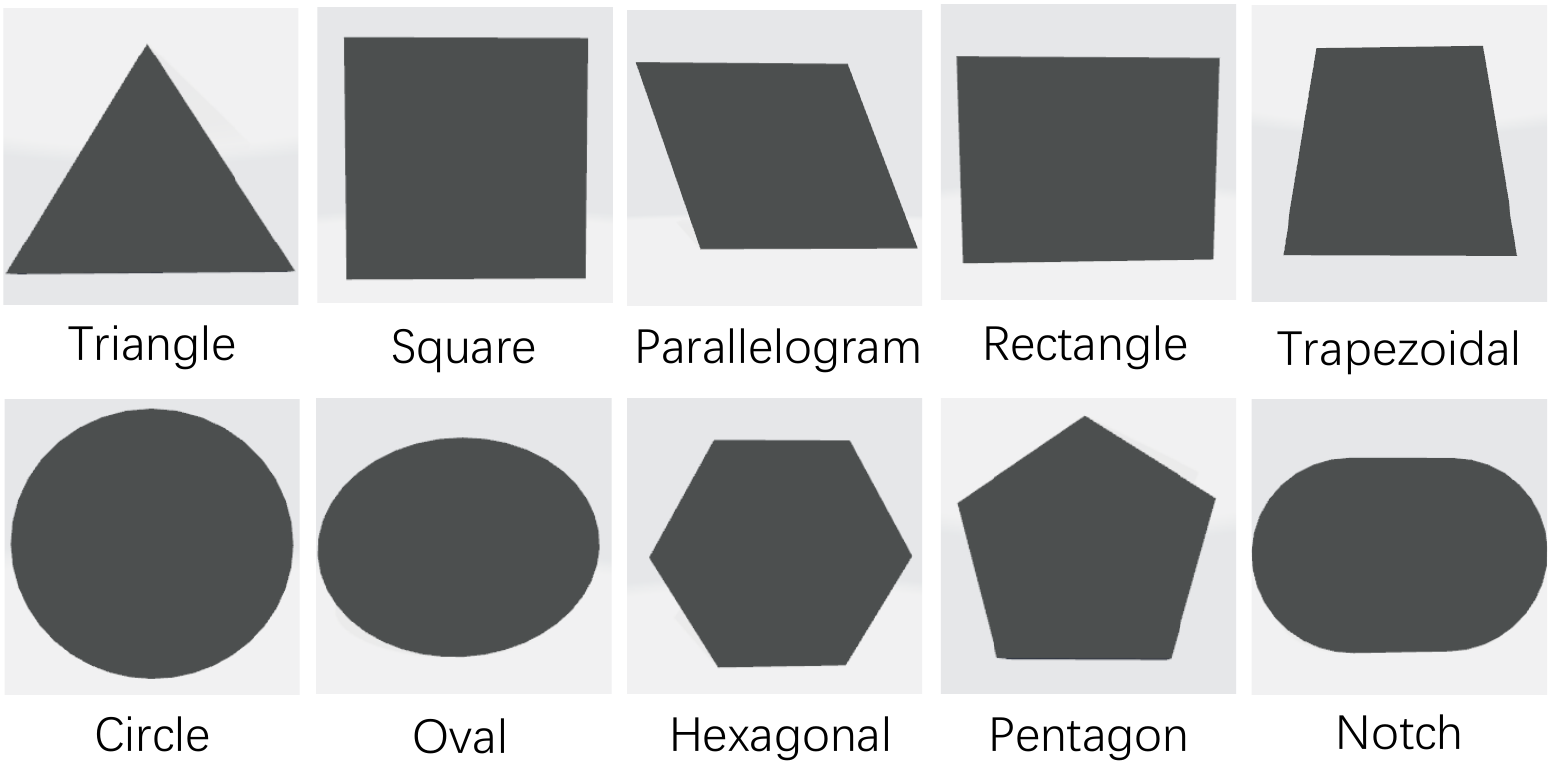}
	\caption{The object dataset used for training consists of ten categories of commonly shaped objects, each containing eight different sizes of objects with inner tangent circle diameters ranging from 165$mm$ to 260$mm$ and height of 60$mm$, with randomly generated colors.}
	\label{Fig. 3}
\end{figure}

\subsection{Sim2Real}
We explore transferring the policy from simulation to a real robot by using CycleGAN~\cite{zhu2017cyclegan} to bridge the gap between simulation and reality. As shown in Fig.~\ref{Fig. 2}, the raw image in the real domain can be transferred to the simulation domain by the generator ($G_{RS}$). We embed the simulated domain image transmitted by $G_{RS}$ into the encoder section to obtain the latent vector, effectively tricking the agent into believing that it is still in the simulator. 

The default CycleGAN network parameters are used in our paper, $Loss$ including $Loss_{GAN}$, $Loss_{cycle}$ and $Loss_{identity}$, and we train it on a dataset of 3000 real images and 3000 simulated images with 180 epochs.

\section{EXPERIMENT}
In this section, we present the experimental setup and the results of the considered manipulation tasks.

\begin{figure}[t] 
	\centering 
	\includegraphics[width=8cm]{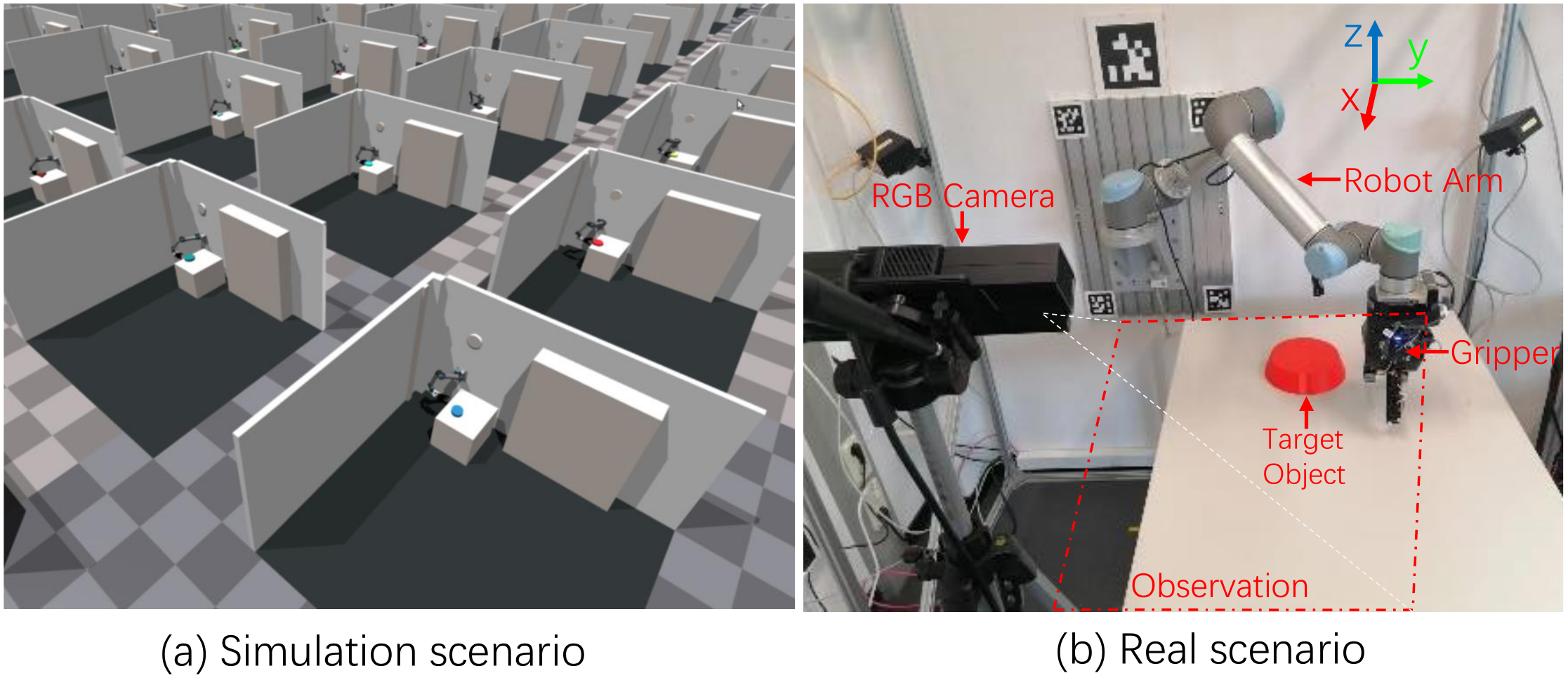} 
	\caption{Simulation and real scenarios. For the simulation scenario, objects are loaded to a random position, and the policy is trained in parallel through 128 environments in Isaac Gym. For the real scenario, the policy is tested on a UR5 robot with a Robotiq 3-finger gripper.}
	\label{Fig. 4}
\end{figure}

In the experiment, a camera captures an RGB image as observation from a static view. The observation includes the gripper, object, table, and floor with an area of $1 m^2$. The training dataset is shown in Fig.~\ref{Fig. 3}. The learning process can be completed within 4 hours in Isaac Gym, an end-to-end GPU-accelerated simulator. For PPO, we use share layers of Actor-Critic, a discount factor $\gamma = 0.99$, a clip $\varepsilon = 0.2$, and the output actions are clipped to range [-1, 1]. The simulated and real-world scenarios are shown in Fig.~\ref{Fig. 4}.

From the experiments, we aim to investigate:
\begin{itemize}
	\item Can one policy network synergize the pushing and grasping actions with higher data efficiency compared to two parallel networks?
	\item Can our policy be applied to objects with beveled surfaces or even some irregular objects with non-overlapping centers of mass and shape?
	\item Can our policy transfer to a real robot without fine-tuning and out-performance push-to-wall baseline?
\end{itemize}

\subsection{Policy Training}
\label{Policy Training}
In previous work, it has been shown to be feasible to train the push and grasp actions separately using two parallel networks~\cite{zeng2018vpg, xu2021grasptarget, yang2021pushandgrasp, berscheid2019shiftandgrasp}. However, using two networks will take longer to train due to the push and grasp data being inefficient to input into the two networks separately and the large size of parameters of the two networks. Moreover, these two actions sometimes do not cooperate well. In our case, we only employ one PPO with shared layers of Actor-Critic, the whole data in pushing and grasping stages input into one network with higher data efficiency (see Sec. \ref{Policy optimization}). For the cooperation of pushing and grasping, a successful grasping needs first pushing the object to a graspable position, so we can achieve well cooperation by separating pushing and grasping stages by time steps (see Sec. \ref{Action}). There are 144 steps in the first stage of pushing, if the gripper pushes the object to the edge before 144 steps, it will learn to stop avoiding pushing the object down. The same is true for the third stage of grasping. If the policy finds a suitable grasping position before total steps, it will stop and wait for the end of this stage.

\subsubsection{\itshape Comparison of using one network and two parallel networks for the PPO policy}
\begin{figure}[t] 
	\centering 
	\includegraphics[width=7cm]{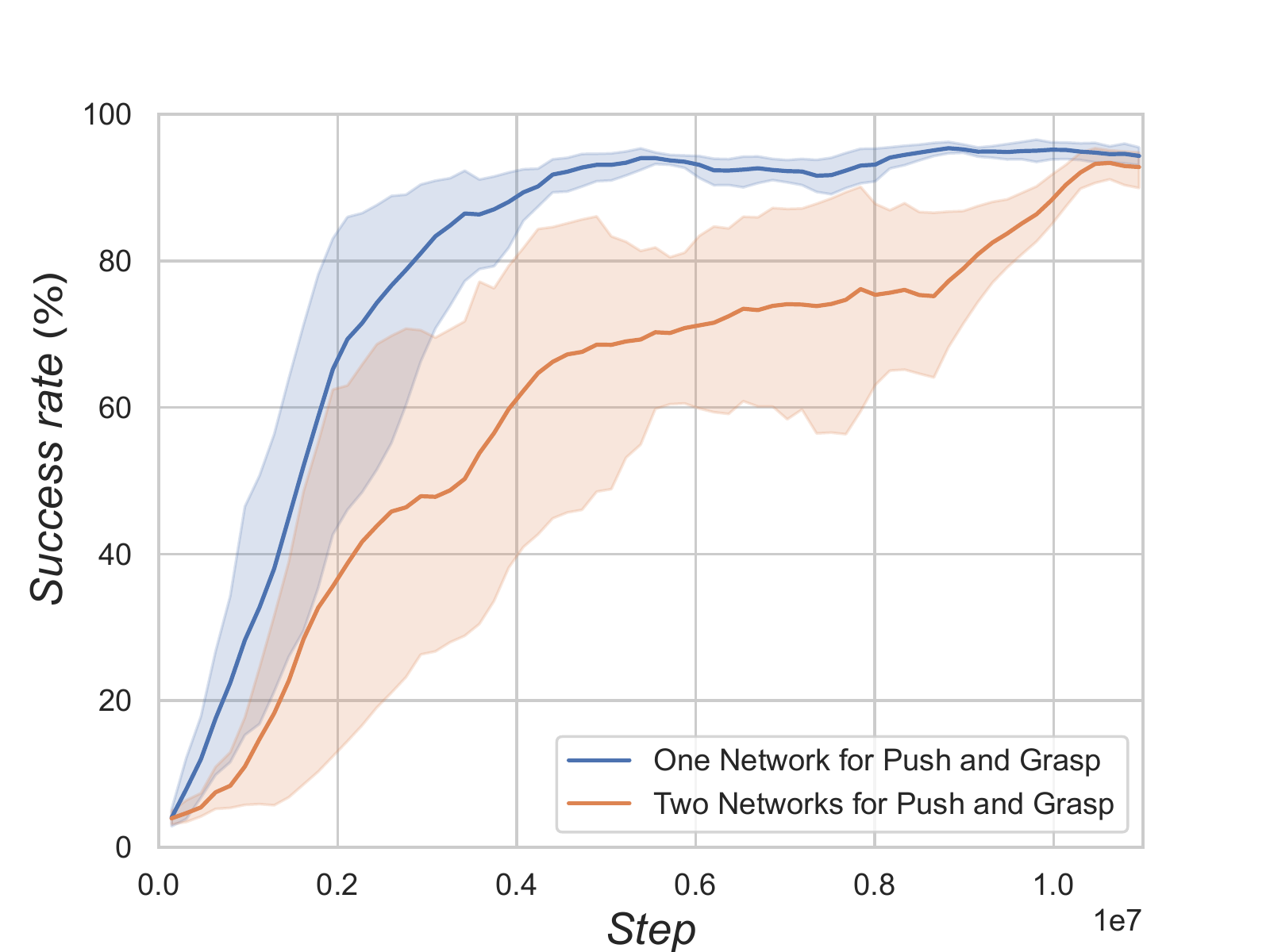}
	\caption{The comparison of training processes between using one network or two networks to learn push and grasp actions.}
	\label{Fig. 5}
\end{figure}

We conducted comparison experiments between one network and two parallel networks learning both push and grasp actions. During all the training processes in this paper, we evaluate the success rate based on 128 environments after every 164k steps, and we consider an object successfully grasped if it is lifted by more than 5~$cm$. The experiments have been run 6 times with different seeds. The results are shown in Fig.~\ref{Fig. 5}.

From the results, we see that the same convergence results can eventually be achieved using a single network or two networks. However, using one network can learn 2.5 times faster meaning it has higher data efficiency. We also see that the variance of one network curve is relatively small, suggesting that diverse data input into a network can make learning more stable.

\subsubsection{\itshape Ablation studies about the common reward and shared layers of Actor-Critic}
\begin{figure}[t]
	\centering
	\includegraphics[width=7.5cm]{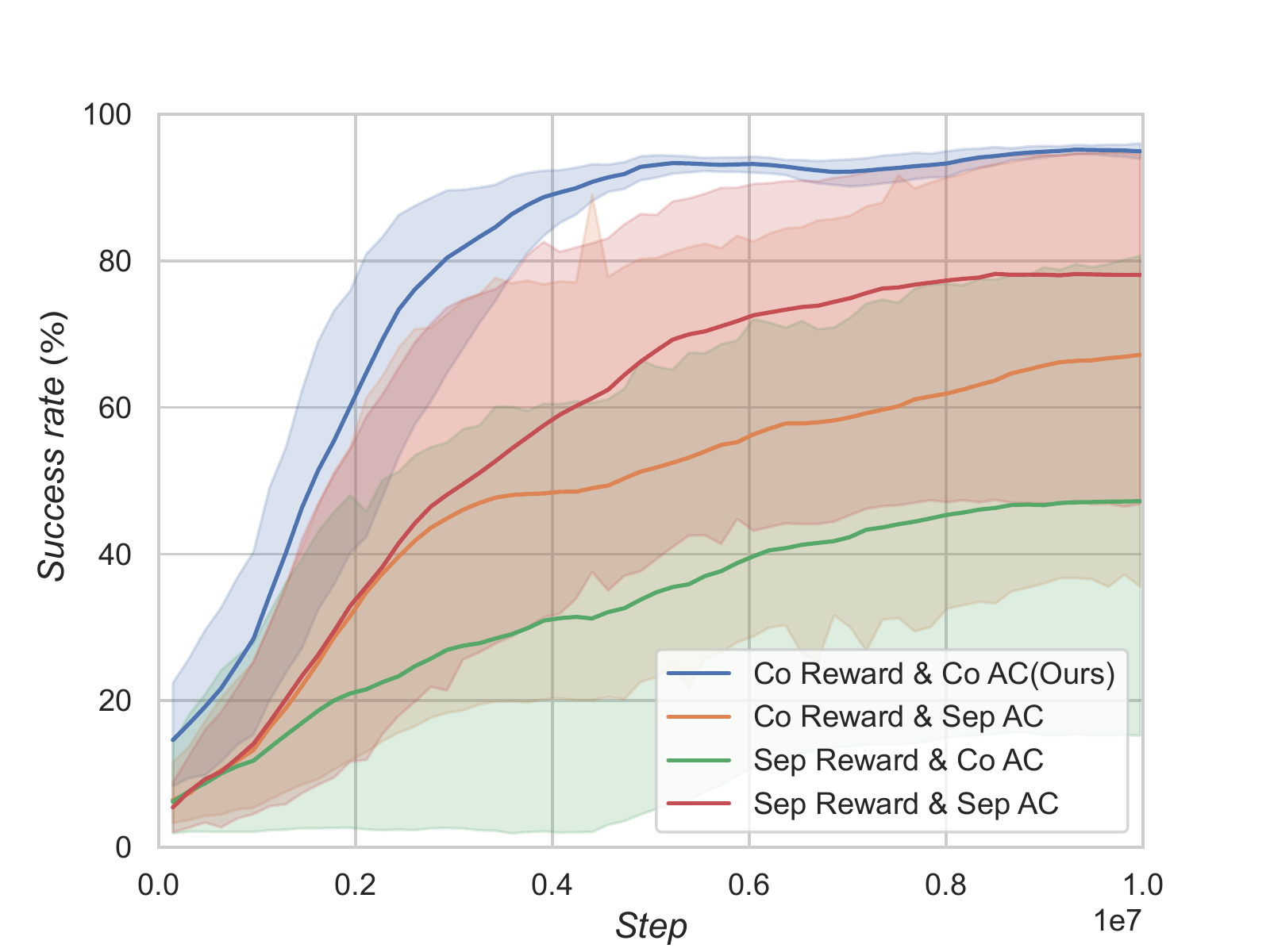}
	\caption{The ablation experiments about whether using a common reward and shared layers of Actor-Critic. ``Co Reward": use a common reward, ``Sep Reward": do not use common reward meaning the rewards in the pushing and the grasping stages are calculated separately, ``Co AC": shared layers of Actor-Critic, ``Sep AC": do not share layers of Actor-Critic.} 
	\label{Fig. 6}
\end{figure}

In our work, the common reward is used to learn common structures between pushing and grasping tasks. With the help of common structures, we can share layers of Actor-Critic to extract the large scale of similar high-dimensional features from multiple parallel environments. The results are shown in Fig.~\ref{Fig. 6}.

From the results, we see that the collaboration of common reward and shared layers of Actor-Critic has the best performance. This verifies that using a common reward can represent common structures between pushing and grasping and the shared layers of Actor-Critic can learn the large scale of common structures better. We also see that if the rewards for the push and grasp stage are calculated separately, the policy tends to learn pushing and grasping as two separate tasks, which means that there are fewer similar structures, and using the shared layers of Actor-Critic makes policy learning worse.

\subsection{Generalization Testing}
Generalization ability is measured by how much variation the trained model can tolerate between the training and testing data. To obtain a more general model, we trained our policy on 80 objects of different sizes, shapes, and random colors (see Fig.~\ref{Fig. 3}). We tested the generalization ability by using objects with similar shapes but with bevels and new objects of irregular shapes.

\subsubsection{\itshape Objects with bevels}
Grasping large flat objects with bevels is a challenging task. The previous methods~\cite{liang2021shovel_and_grasp, sun2020push_to_wall} can not handle these objects because there is not enough space for the gripper to push it against a wall to the hanging state. However, in our strategy, we push the object to the edge of the table to create the hanging part which is less sensitive to the shape of the side of the objects.

We created the test dataset with the similar shape as the training set, and each category has one object with the random value of the inner tangent circle diameter from 165$mm$ to 260$mm$, was stretched by 50$mm$ in the axial direction, object with a chamfer of $30^{\circ}$, and random color. We tested the ten category objects in the best-performing model in four ablation strategies to compare the generalization performance. The results are shown in Table~\ref{tab:1}. 

We observe that our method (Co Reward \& Co AC) has the best performance, achieving more than 80\% success rate for all beveled objects, and the rounder the object, the better the performance. The results imply that our model can be generalized to beveled objects. We also see that the common reward models perform better than the separate reward models, meaning the common reward plays a vital role in bevel object scenarios.

\begin{table*}[htbp]
	\centering
	\setlength{\tabcolsep}{4.7pt}
	\caption{Generalization test for beveled objects (mean \%)}
	\label{tab:1}
	\begin{threeparttable}  
	\begin{tabular}{cccccccccccc}
		\toprule
		Shape & Triangles & Square & Parallelogram & Rectangle & Trapezoidal & Circle & Oval & Hexagonal & Pentagon & Notch & All \\
		\midrule
		Sep Reward \& Sep AC         & 55.5 & 62.5 & 58.8 & 54.2 & 51.0 & 64.2 & 67.2 & 51.2 & 63.0 & 69.7 & 59.2 \\
		Sep Reward \& Co AC          & 34.0 & 40.7 & 40.2 & 38.3 & 38.3 & 44.8 & 43.8 & 36.5 & 42.8 & 43.2 & 40.8 \\
		Co Reward \& Sep AC          & 63.7 & 77.0 & 69.7 & 70.0 & 74.8 & 83.3 & 82.3 & 76.3 & 79.7 & 81.3 & 74.2 \\
		Co Reward \& Co AC (Ours)    & \textbf{80.2} & \textbf{91.2} & \textbf{86.2} & \textbf{88.2} & \textbf{86.3} & \textbf{97.2} & \textbf{95.5} & \textbf{89.2} & \textbf{93.0} & \textbf{91.3} & \textbf{86.2} \\
		\bottomrule
	\end{tabular}
	\end{threeparttable}
\end{table*}

\begin{table*}[htbp]
	\centering
	\setlength{\tabcolsep}{5pt}
	\caption{Generalization test for irregularly shaped objects (mean \%)}
	\label{tab:2}
	\begin{threeparttable}  
		\begin{tabular}{cp{1.12cm}<{\centering}cp{1.12cm}<{\centering}cp{1.12cm}<{\centering}cp{1.12cm}<{\centering}cp{1.12cm}<{\centering}cp{1.12cm}<{\centering}cp{1.12cm}<{\centering}cp{1.12cm}<{\centering}cp{1.12cm}<{\centering}cp{1.12cm}<{\centering}cp{1.12cm}<{\centering}cp{1.12cm}<{\centering}}
			\toprule
			Object ID & 1 & 2 & 3 & 4 & 5 & 6 & 7 & 8 & 9 & 10 & 11 & All \\
			\midrule
			Sep Reward \& Sep AC         & 74.3 & 78.3 & 73.3 & 83 & 70.3 & 55.7 & 83.0 & 77.7 & 72.3 & 68.3 & 83.0 & 75.5 \\
			Sep Reward \& Co AC          & 44.5 & 49.0 & 49.3 & 50.2 & 42.8 & 32.7 & 49.7 & 42.3 & 46.5 & 44.7 & 50.0 & 47.3 \\
			Co Reward \& Sep AC          & 76.2 & 75.5 & 73.8 & 83.8 & 73.8 & 63.5 & 83.8 & 76.3 & 69.8 & 67.3 & 84.3 & 77.7 \\
			Co Reward \& Co AC (Ours)    & \textbf{82.7} & \textbf{91.3} & \textbf{93.0} & \textbf{100.0} & \textbf{84.8} & \textbf{75.8} & \textbf{98.8} & \textbf{83.8} & \textbf{80.5} & \textbf{80.8} & \textbf{100.0} & \textbf{90.3} \\
			\bottomrule
		\end{tabular}
	\end{threeparttable}
\end{table*}

\begin{table*}[htbp]
	\centering
	\setlength{\tabcolsep}{5pt}
	\caption{Real robot test (mean \%)}
	\label{tab:3}
	\begin{threeparttable}  
		\begin{tabular}{cp{0.9cm}<{\centering}cp{0.9cm}<{\centering}cp{0.9cm}<{\centering}cp{0.9cm}<{\centering}cp{0.9cm}<{\centering}cp{0.9cm}<{\centering}cp{0.9cm}<{\centering}cp{0.9cm}<{\centering}cp{0.9cm}<{\centering}cp{0.9cm}<{\centering}cp{0.9cm}<{\centering}cp{0.9cm}<{\centering}cp{0.9cm}<{\centering}}
			\toprule
			Object ID & 1 & 2 & 3 & 4 & 5 & 6 & 7 & 8 & 9 & 10 & 11 & 12 & 13 \\
			\midrule
			Baseline         & 40.0 & 33.3 & 26.7 & 46.7 & 0.0 & 13.3 & 40.0 & \textbf{86.7} & 0.0 & 0.0 & 0.0 & 0.0 & 0.0 \\
			Co Reward \& Co AC (Ours)    & \textbf{86.7} & \textbf{73.3} & \textbf{86.7} & \textbf{66.7} & \textbf{80.0} & \textbf{86.7} & \textbf{73.3} & 73.3 & \textbf{80.0} & \textbf{80.0} & \textbf{80.0} & \textbf{73.3} & \textbf{93.3} \\
			\bottomrule
		\end{tabular}
	\end{threeparttable}
\end{table*}

\subsubsection{\itshape Irregularly shaped objects}
\begin{figure}[t]
	\centering 
	\includegraphics[width=8cm]{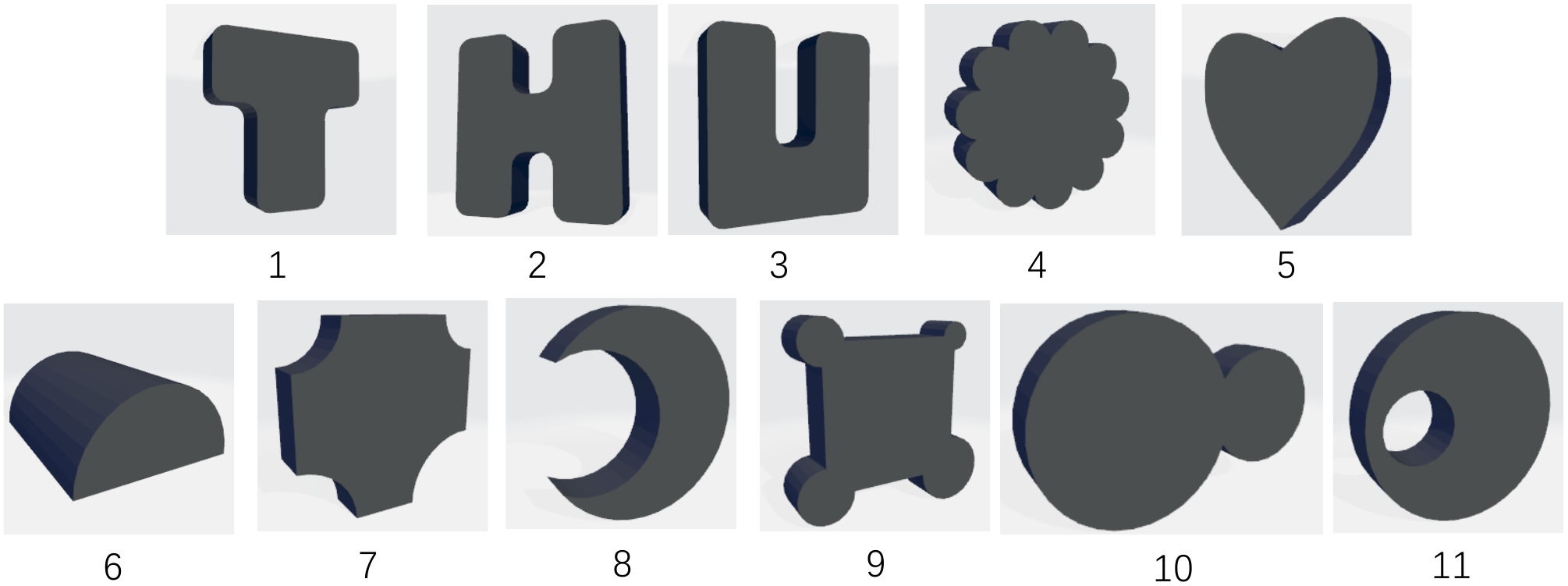}
	\caption{The irregularly shaped objects dataset used for testing consists of letters and large-scale beveled objects, most of them with non-overlapping centers of mass and shape. The height of the objects is 60$mm$ and the colors are randomly generated.}
	\label{Fig. 7}
\end{figure}

In addition to testing similar objects with bevels, we also built new, more challenging, irregularly shaped objects to test the generalization ability of our model. The challenging dataset is shown in Fig.~\ref{Fig. 7}. For these irregularly shaped objects, if the heavier part hangs on the edge of the table, it is easier to fall off. The results are shown in Table~\ref{tab:2}.

In terms of the results, it still maintains a high success rate. We can see that regardless of the shapes of objects, our policy tends to push a small part of the object out of the table, reflecting a good trade-off between pushing the object to an easily graspable position and avoiding falling. Comparing Table~\ref{tab:1} and Table~\ref{tab:2}, the performance of irregular objects is slightly better than that of objects with bevels because the bevels will change the movement of the gripper, causing an error between the target position and the actual position, which also indicates that if the gripper can execute actions accurately, the policy can be adapted to new objects, even though the object shape may be very irregular.

\subsection{Real Robot Experiments}
\begin{figure}[t]
	\centering
	\includegraphics[width=7.5cm]{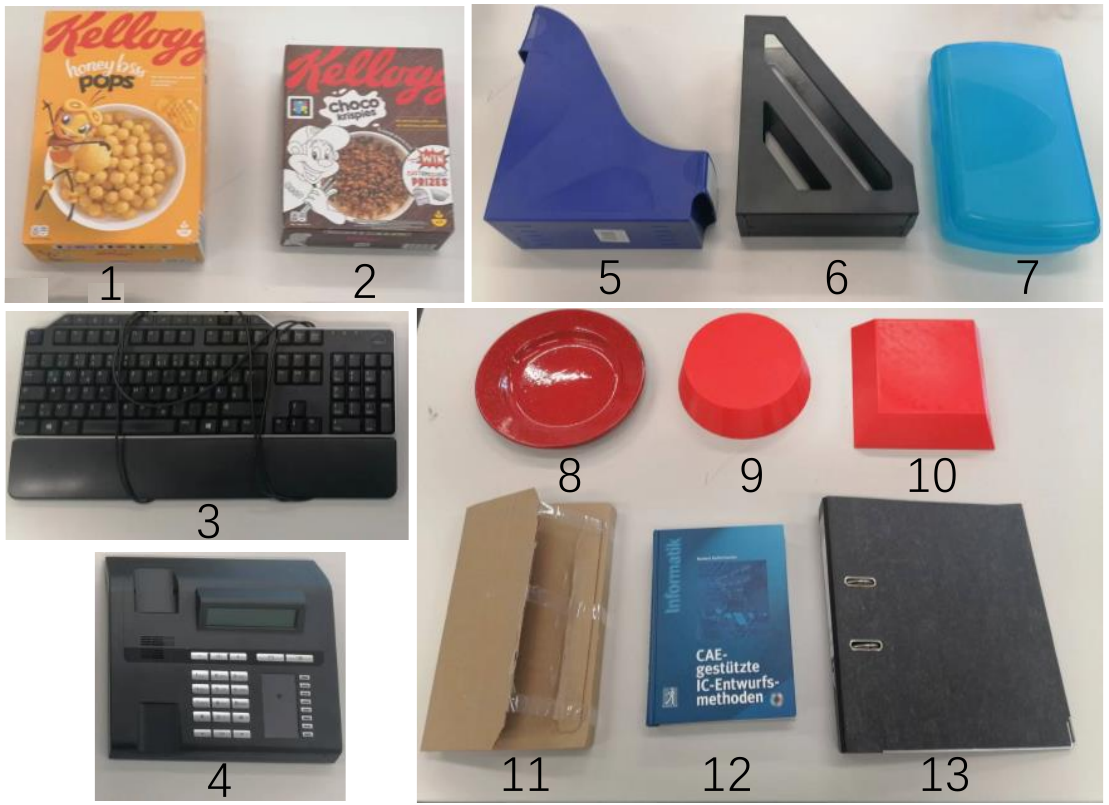}
	\caption{The real robot experiments dataset includes different kinds of daily objects. Objects range in width from 180$mm$ to 440$mm$, and in height from 25$mm$ to 95$mm$.} 
	\label{Fig. 8}
\end{figure}

For the real robot experiments, we used 13 daily large flat objects as the dataset (see Fig.~\ref{Fig. 8}). The experimental scenario is shown in Fig.~\ref{Fig. 4}(b). To deal with the gap between the simulated and real images, we employ CycleGAN to transfer images from the real domain to the simulated domain and then embed them into the trained policy. We consider a push-to-wall method~\cite{liang2021shovel_and_grasp} as a baseline, but different with~\cite{liang2021shovel_and_grasp} we let the gripper push against a wall along the center of the object instead of learning a push point to achieve better performances. We compared the success rate of the baseline and our method through 15 attempts, and the results are shown in Table~\ref{tab:3}.

From the results, we observe that our model is transferable to a real robot that can successfully grasp daily objects without fine-tuning. Focusing on the baseline, we observe that the method of push-to-wall can not handle objects with bevels or with hard flat sides. Our model is less sensitive to object shapes and materials, moreover, we can achieve respond dynamically to the disturbance of object position changes. However, there are some failures i.e. the robot pushes too much to make the object fall and the success rate is lower than the simulated experiments, these are because real objects always have complex textures, which is a big challenge for our visual-base model. More details can be found in \url{https://haozhang990127.github.io/PaG/}.

\section{CONCLUSIONS}
In this paper, we proposed a challenging visual-based push and grasp manipulation for objects from ungraspable poses that can only be successfully grasped by first pushing the object to the edge of the table. To solve this task with high data efficiency, we employ one PPO network with shared layers of Actor-Critic to learn both pushing and grasping actions. Our extensive experiments show that our PaG method can effectively train a policy to adapt to different objects, even with bevels or irregular shapes. In addition, by using CycleGAN for domain adaption, the trained policy can be transferred to the real world without fine-tuning.

In our method, we only considered the uniform density of an ungraspable object. Training the robot to push and grasp ungraspable objects with uneven density, and identifying the center of mass from the trajectory is meaningful. Stacking scenarios with multiple graspable and ungraspable objects is also a challenging task for further exploration.

\bibliographystyle{IEEEtran}
\bibliography{mybib}

\begin{thebibliography}{10}
\providecommand{\url}[1]{#1}
\csname url@samestyle\endcsname
\providecommand{\newblock}{\relax}
\providecommand{\bibinfo}[2]{#2}
\providecommand{\BIBentrySTDinterwordspacing}{\spaceskip=0pt\relax}
\providecommand{\BIBentryALTinterwordstretchfactor}{4}
\providecommand{\BIBentryALTinterwordspacing}{\spaceskip=\fontdimen2\font plus
\BIBentryALTinterwordstretchfactor\fontdimen3\font minus
  \fontdimen4\font\relax}
\providecommand{\BIBforeignlanguage}[2]{{%
\expandafter\ifx\csname l@#1\endcsname\relax
\typeout{** WARNING: IEEEtran.bst: No hyphenation pattern has been}%
\typeout{** loaded for the language `#1'. Using the pattern for}%
\typeout{** the default language instead.}%
\else
\language=\csname l@#1\endcsname
\fi
#2}}
\providecommand{\BIBdecl}{\relax}
\BIBdecl

\bibitem{liang2021shovel_and_grasp}
H.~Liang, X.~Lou, Y.~Yang, and C.~Choi, ``Learning visual affordances with
  target-orientated deep q-network to grasp objects by harnessing environmental
  fixtures,'' in \emph{IEEE International Conference on Robotics and Automation
  (ICRA)}.\hskip 1em plus 0.5em minus 0.4em\relax IEEE, 2021, pp. 2562--2568.

\bibitem{sun2020push_to_wall}
Z.~Sun, K.~Yuan, W.~Hu, C.~Yang, and Z.~Li, ``Learning pregrasp manipulation of
  objects from ungraspable poses,'' in \emph{IEEE International Conference on
  Robotics and Automation (ICRA)}.\hskip 1em plus 0.5em minus 0.4em\relax IEEE,
  2020, pp. 9917--9923.

\bibitem{zeng2018vpg}
A.~Zeng, S.~Song, S.~Welker, J.~Lee, A.~Rodriguez, and T.~Funkhouser,
  ``Learning synergies between pushing and grasping with self-supervised deep
  reinforcement learning,'' in \emph{IEEE/RSJ International Conference on
  Intelligent Robots and Systems (IROS)}.\hskip 1em plus 0.5em minus
  0.4em\relax IEEE, 2018, pp. 4238--4245.

\bibitem{mnih2013dqn}
V.~Mnih, K.~Kavukcuoglu, D.~Silver, A.~Graves, I.~Antonoglou, D.~Wierstra, and
  M.~Riedmiller, ``Playing atari with deep reinforcement learning,''
  \emph{arXiv preprint arXiv:1312.5602}, 2013.

\bibitem{kingma2013vae}
D.~P. Kingma and M.~Welling, ``Auto-encoding variational bayes,'' \emph{arXiv
  preprint arXiv:1312.6114}, 2013.

\bibitem{schulman2017ppo}
J.~Schulman, F.~Wolski, P.~Dhariwal, A.~Radford, and O.~Klimov, ``Proximal
  policy optimization algorithms,'' \emph{arXiv preprint arXiv:1707.06347},
  2017.

\bibitem{xu2021grasptarget}
K.~Xu, H.~Yu, Q.~Lai, Y.~Wang, and R.~Xiong, ``Efficient learning of
  goal-oriented push-grasping synergy in clutter,'' \emph{IEEE Robotics and
  Automation Letters}, vol.~6, no.~4, pp. 6337--6344, 2021.

\bibitem{cong2022reinforcement}
L.~Cong, H.~Liang, P.~Ruppel, Y.~Shi, M.~G{\"o}rner, N.~Hendrich, and J.~Zhang,
  ``Reinforcement learning with vision-proprioception model for robot planar
  pushing,'' \emph{Frontiers in Neurorobotics}, vol.~16, 2022.

\bibitem{haarnoja2018sac}
T.~Haarnoja, A.~Zhou, P.~Abbeel, and S.~Levine, ``Soft actor-critic: Off-policy
  maximum entropy deep reinforcement learning with a stochastic actor,'' in
  \emph{International conference on machine learning}.\hskip 1em plus 0.5em
  minus 0.4em\relax PMLR, 2018, pp. 1861--1870.

\bibitem{breyer2019closeloopgrasp}
M.~Breyer, F.~Furrer, T.~Novkovic, R.~Siegwart, and J.~Nieto, ``Comparing task
  simplifications to learn closed-loop object picking using deep reinforcement
  learning,'' \emph{IEEE Robotics and Automation Letters}, vol.~4, no.~2, pp.
  1549--1556, 2019.

\bibitem{schulman2015trpo}
J.~Schulman, S.~Levine, P.~Abbeel, M.~Jordan, and P.~Moritz, ``Trust region
  policy optimization,'' in \emph{International conference on machine
  learning}, 2015, pp. 1889--1897.

\bibitem{yang2021pushandgrasp}
Y.~Yang, Z.~Ni, M.~Gao, J.~Zhang, and D.~Tao, ``Collaborative pushing and
  grasping of tightly stacked objects via deep reinforcement learning,''
  \emph{IEEE/CAA Journal of Automatica Sinica}, vol.~9, no.~1, pp. 135--145,
  2021.

\bibitem{coumans2016pybullet}
E.~Coumans and Y.~Bai, ``Pybullet, a python module for physics simulation for
  games, robotics and machine learning,'' \emph{GitHub repository}, 2016.

\bibitem{todorov2012mujoco}
E.~Todorov, T.~Erez, and Y.~Tassa, ``Mujoco: A physics engine for model-based
  control,'' in \emph{IEEE/RSJ international conference on intelligent robots
  and systems}.\hskip 1em plus 0.5em minus 0.4em\relax IEEE, 2012, pp.
  5026--5033.

\bibitem{makoviychuk2021isaacgym}
V.~Makoviychuk, L.~Wawrzyniak, Y.~Guo, M.~Lu, K.~Storey, M.~Macklin,
  D.~Hoeller, N.~Rudin, A.~Allshire, A.~Handa \emph{et~al.}, ``Isaac gym: High
  performance gpu based physics simulation for robot learning,'' in
  \emph{Thirty-fifth Conference on Neural Information Processing Systems
  Datasets and Benchmarks Track (Round 2)}, 2021.

\bibitem{tobin2017sim2real}
J.~Tobin, R.~Fong, A.~Ray, J.~Schneider, W.~Zaremba, and P.~Abbeel, ``Domain
  randomization for transferring deep neural networks from simulation to the
  real world,'' in \emph{IEEE/RSJ international conference on intelligent
  robots and systems (IROS)}.\hskip 1em plus 0.5em minus 0.4em\relax IEEE,
  2017, pp. 23--30.

\bibitem{james2017sim2real}
S.~James, A.~J. Davison, and E.~Johns, ``Transferring end-to-end visuomotor
  control from simulation to real world for a multi-stage task,'' in
  \emph{Conference on Robot Learning}, 2017, pp. 334--343.

\bibitem{tan2018sim2real}
J.~Tan, T.~Zhang, E.~Coumans, A.~Iscen, Y.~Bai, D.~Hafner, S.~Bohez, and
  V.~Vanhoucke, ``Sim-to-real: Learning agile locomotion for quadruped
  robots,'' \emph{arXiv preprint arXiv:1804.10332}, 2018.

\bibitem{zhu2017cyclegan}
J.-Y. Zhu, T.~Park, P.~Isola, and A.~A. Efros, ``Unpaired image-to-image
  translation using cycle-consistent adversarial networks,'' in
  \emph{Proceedings of the IEEE international conference on computer vision},
  2017, pp. 2223--2232.

\bibitem{kingma2014adam}
D.~P. Kingma and J.~Ba, ``Adam: A method for stochastic optimization,''
  \emph{arXiv preprint arXiv:1412.6980}, 2014.

\bibitem{yu2020meta}
T.~Yu, D.~Quillen, Z.~He, R.~Julian, K.~Hausman, C.~Finn, and S.~Levine,
  ``Meta-world: A benchmark and evaluation for multi-task and meta
  reinforcement learning,'' in \emph{Conference on robot learning}, 2020, pp.
  1094--1100.

\bibitem{mnih2016a3c}
V.~Mnih, A.~P. Badia, M.~Mirza, A.~Graves, T.~Lillicrap, T.~Harley, D.~Silver,
  and K.~Kavukcuoglu, ``Asynchronous methods for deep reinforcement learning,''
  in \emph{International conference on machine learning}.\hskip 1em plus 0.5em
  minus 0.4em\relax PMLR, 2016, pp. 1928--1937.

\bibitem{shukla2021shareac}
P.~Shukla, M.~Pegu, and G.~Nandi, ``Development of behavior based robot
  manipulation using actor-critic architecture,'' in \emph{2021 8th
  International Conference on Signal Processing and Integrated Networks
  (SPIN)}.\hskip 1em plus 0.5em minus 0.4em\relax IEEE, 2021, pp. 469--474.

\bibitem{berscheid2019shiftandgrasp}
L.~Berscheid, P.~Mei{\ss}ner, and T.~Kr{\"o}ger, ``Robot learning of shifting
  objects for grasping in cluttered environments,'' in \emph{IEEE/RSJ
  International Conference on Intelligent Robots and Systems (IROS)}.\hskip 1em
  plus 0.5em minus 0.4em\relax IEEE, 2019, pp. 612--618.

\end{thebibliography}
\end{document}